\documentclass[final]{cvpr}

\usepackage{cite}

\usepackage[margin=6pt,font=small,labelfont=bf,labelsep=endash,tableposition=top]{caption}

\usepackage{epsfig}
\usepackage{graphicx}
\usepackage{amsmath}
\usepackage{amssymb}
\usepackage{bbm,nicefrac}
\usepackage{makecell, verbatim, sidecap}
\usepackage[position=bottom]{subfig}

\usepackage[pagebackref=true,breaklinks=true,colorlinks,bookmarks=false]{hyperref}

\def\R{{\mathbb R}}
\def\max{{\rm max}}
\def\Proj{{\rm Proj}}

\usepackage{multirow, xspace}

\renewcommand{\vec}[1]{\ensuremath{\pmb{#1}}}
\newcommand{\mat}[1]{\ensuremath{\mathbf{#1}}}
\newcommand{\set}[1]{\ensuremath{\mathscr{#1}}}

\makeatletter
\@tfor\next:=abcdefghijklmnopqrstuvwxyxz\do
{\begingroup\edef\x{\endgroup
        \noexpand\@namedef{v\next}{\noexpand\vec{\next}}%
    }\x}
\@tfor\next:=ABCDEFGHIJKLMNOPQRSTUVWXYZ\do
{\begingroup\edef\x{\endgroup
        \noexpand\@namedef{m\next}{\noexpand\mat{\next}}%
    }\x}
\@tfor\next:=ABCDEFGHIJKLMNOPQRSTUVWXYZ\do
{\begingroup\edef\x{\endgroup
        \noexpand\@namedef{s\next}{\noexpand\set{\next}}%
    }\x}
\makeatother

\def\Ours{{BoxInst}\xspace}

\begin{document}

\title{\Ours:
High-Performance Instance Segmentation with Box Annotations
}

\author{Zhi Tian, ~ ~ ~ Chunhua Shen\thanks{Corresponding author.},
~  ~ ~ Xinlong Wang, ~  ~ ~ Hao Chen
\\[0.25cm]
The University of Adelaide, Australia
}

\makeatletter
\let\@oldmaketitle\@maketitle
\renewcommand{\@maketitle}{\@oldmaketitle
 \centering
    \includegraphics[trim=0 .25cm 0 0,clip,width=\linewidth]{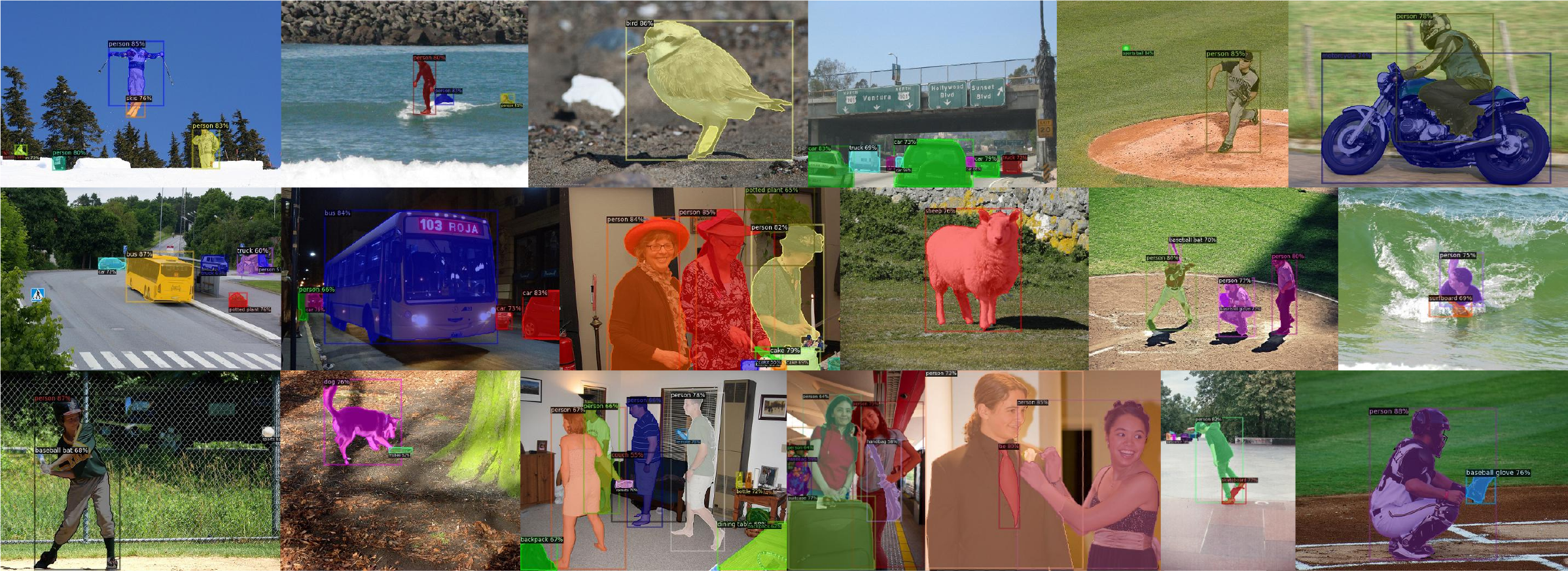}
    \captionof{figure}{
\textbf{Some qualitative results of \Ours}  with the ResNet-101 based model achieving 33.0\% mask AP on COCO \texttt{val2017}. The model is trained without any mask annotations and can infer at 10 FPS on a 1080Ti GPU. Best viewed on screen.
    }
    \label{fig:vis_r101_3x}
    \bigskip}                   
\makeatother
\maketitle

\maketitle

\begin{abstract}

We present a high-performance method that can achieve mask-level instance segmentation with only bounding-box annotations for training. While this setting has been studied in the literature, here we show significantly stronger performance with a simple design (e.g., {dramatically improving previous best reported mask AP of {\bf 21.1\%} \cite{hsu2019weakly} {\bf  to 31.6\% } on the COCO dataset}). Our core idea is to redesign the loss of learning masks in instance segmentation, with no modification to the segmentation network itself. The new loss functions can supervise the mask training without relying on mask annotations. This is made possible with two loss terms, namely, 1) a surrogate term that minimizes the discrepancy between the projections of the ground-truth box and the predicted mask; 2) a pairwise loss that can exploit the prior that proximal pixels with similar colors are very likely to have the same category label.

Experiments demonstrate that the redesigned mask loss can yield surprisingly high-quality instance masks with only box annotations. For example, without using any mask annotations, with a ResNet-101 backbone and 3$\times$ training schedule, we achieve 33.2\% mask AP on COCO test-dev split (vs.\ 39.1\% of the fully supervised counterpart). Our excellent experiment results on COCO and Pascal VOC indicate that our method dramatically narrows the performance gap between weakly and fully supervised instance segmentation.

{   \def\UrlFont{\sf}
    \def\UrlFont{\rm\small\ttfamily}
Code is available at: \url{https://git.io/AdelaiDet}
}

\end{abstract}

\section{Introduction}

\begin{figure}[t!]
\centering
\includegraphics[width=\linewidth]{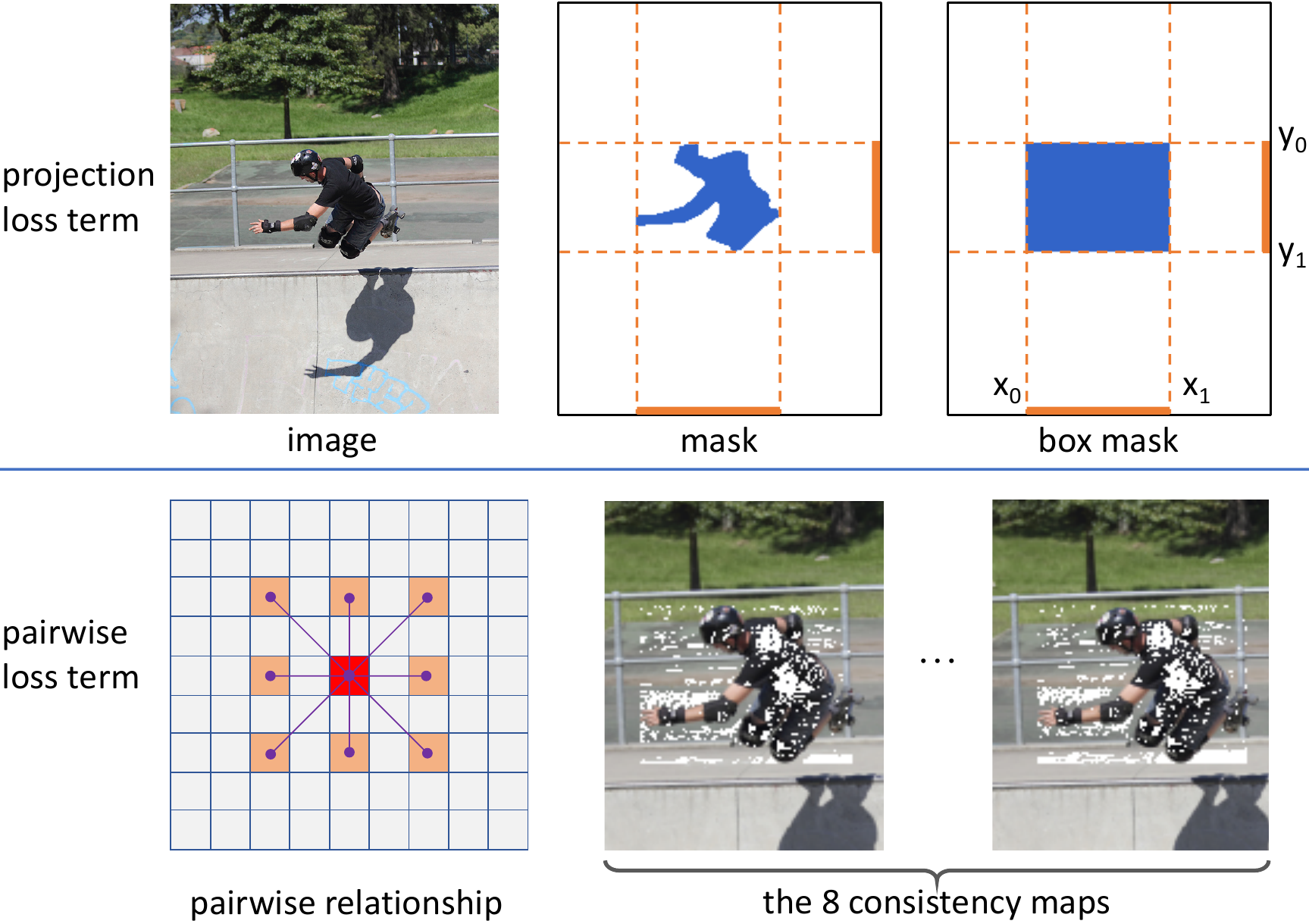}
   \caption{
   \textbf{The  two proposed  loss terms}.
   Top row: the projections onto $x$-axis and $y$-axis of the mask and the box, and the projections should be the same, where $(x_0, y_0)$ and $(x_1, y_1)$ are the two corners of the box.
   Bottom row:
   the pairwise term. For each pixel, we compute the pairwise label consistency between the pixel and its 8 neighbours (with dilation rate 2). Thus each pixel has 8 edges and we have 8 consistency maps in the right. The white locations in the right figure are the edges we have the supervision derived from the color similarity, and other edges are discarded in the loss computation.}
\label{fig:vis_loss_terms}

\end{figure}

Instance segmentation requires the algorithm to predict the
pixel-wise
masks and categories of instances of interest, and is %
one of the most fundamental tasks in computer vision. The performance of instance segmentation has been significantly advanced by a number of recent successful methods~\cite{he2017mask, huang2019mask, tian2020conditional, SoloV22020, wang2020solo, ChengWHL20, chen2020blendmask}. These methods have almost made the previously much more challenging instance segmentation task be as simple and fast as bounding-box object detection. For example, built on the detector FCOS~\cite{tian2019fcos}, CondInst~\cite{tian2020conditional} only adds very compact dynamic mask heads to
predict instance masks,
and thus only %
introduces less than 10\% computation time, compared to FCOS.
Instance segmentation is able to provide more accurate and finer
\textit{mask-level} object location
than detection. Thus, given that the extra computation cost is negligible, instance segmentation should be preferred over bounding box detection in many cases. For example, if a robot wants to grasp an object, an accurate mask will be much more helpful than a box.
Now the main obstacle that impedes instance segmentation replacing box detection
is the significantly heavier pixel-wise mask annotations.
Compared to box-level annotations required by object detection, annotating
pixel-level masks is notoriously time-consuming, as shown in \cite{bearman2016s, everingham2010Pascal, kulharia12356box2seg}.
Here we aim to eliminate this obstacle by training instance segmentation using box annotations only.

A few
works \cite{song2019box, dai2015boxsup, kulharia12356box2seg, papandreou2015weakly, hsu2019weakly, khoreva2017simple, rajchl2016deepcut, arun2020weakly}
attempted
to obtain (semantic     or instance-level) mask prediction
with box-level annotations. Among them, most methods such as BoxSup
\cite{dai2015boxsup} and Box2Seg~\cite{kulharia12356box2seg} rely on the region proposals that are generated by MCG~\cite{pont2016multiscale} or GrabCut~\cite{rother2004grabcut}.
One drawback might be the slow
training procedure
since these algorithms are hard to be parallelized by modern GPUs.
Moreover, in order to achieve good performance, some %
methods often require iterative training, resulting in a complicated training pipeline and more hyper-parameters. Most importantly, \textit{none of these methods is able to show  strong weakly-supervised performance on large
benchmarks such as COCO~\cite{lin2014microsoft}.} %
Thus almost all of them
are only evaluated on %
small datasets such as Pascal VOC~\cite{everingham2010Pascal}.

In this work, we propose a simple, single-shot and high-performance box-supervised instance segmentation method, built upon the recent fully convolutional instance segmentation framework---CondInst~\cite{tian2020conditional}. Our core idea is to replace the original pixel-wise mask losses in CondInst with a %
carefully designed
mask loss consisting of two terms.  The first term
minimizes the discrepancy between the horizontal and vertical projections of the predicted mask and the ground-truth box (see Fig.~\ref{fig:vis_loss_terms} top). \textit{This essentially ensures that the tightest box covering the predicted mask matches the ground-truth box.} Since the ground-truth mask and ground-truth box have the same projections on the two axes\footnote{This may not hold if the instance mask consists multiple disjointed regions.}, this can be also viewed as a surrogate term that minimizes the discrepancy between the projections of the predicted mask and ground-truth mask.
This loss term %
can be computed
when we only have box annotations.
Clearly, with this projection term, multiple masks can be projected to
a same box. Therefore the projection loss alone would not suffice.
Thus, we introduce
the second loss term,  encouraging the prediction and ground-truth masks have the same \textit{pairwise label similarity} in proximal pixels (Fig.~\ref{fig:vis_loss_terms} bottom).
At first glance, the pairwise similarity of the ground-truth masks cannot be computed if we do not have the mask annotations. With only box annotations available,  in principle this pairwise supervision signal is inevitably noisy.
However, \textit{an important observation is that the proximal pixels with similar colors are very likely to have the same label.}
Thus, we show that it is
empirically plausible to determine a color similarity threshold such that
only confident pairs of pixels having a same label are used in the loss computation (the white regions in the bottom right of Fig.~\ref{fig:vis_loss_terms}), thus largely eliminating supervision noises.
Using
these two loss terms, we achieve stunning instance segmentation results \textit{without using any mask annotations}. Some qualitative results are shown in Fig.~\ref{fig:vis_r101_3x}.
Even though ideas that are relevant  to  either of our two observations mentioned above were studied more or less in the literature,
ranging from non-deep learning methods such as CRF~\cite{krahenbuhl2011efficient} and GrabCut~\cite{rother2004grabcut} to deep learning-based methods such as Box2Seg~\cite{kulharia12356box2seg} and BBTP~\cite{hsu2019weakly},
none of these works effectively  incorporates them into a simple and appropriate
framework.
As a result, and more importantly,
performance of existing methods on large challenging datasets (\textit{e.g.},
COCO) is far away from that of the full potential of box-supervised instance segmentation that is achievable, as we are going to reveal here.
In summary,
our method, termed {\bf \Ours},  enjoys the following advantages.
\begin{itemize}
\itemsep -0.152cm
    \item The proposed  method
    can %
    achieve
    instance segmentation %
    with box supervision by %
    introducing
    two loss terms to the
    instance segmentation
    framework CondInst~\cite{tian2020conditional}.
    \Ours\ is \textit{simple} as it does not
    modify the network model of CondInst at all, only using different loss terms. This means that  the inference process of the proposed \Ours\ is exactly the same as CondInst, thus naturally inheriting
    all  desirable properties of CondInst.

    \item \Ours\ attains excellent instance segmentation performance on the large-scale benchmark COCO.
    With the ResNet-101 backbone and 3$\times$ training schedule, our \Ours\ achieves
    33.2\% mask AP on COCO with no mask annotations used in training,
    \textit{outperforming a few recent fully supervised methods using the same backbone and trained with mask annotations}, including YOLACT \cite{yolact} (31.2\% AP) and PolarMask \cite{xie2020polarmask} (32.1\% AP). We empirically show that in the semi-supervised setting,
    mask AP of \Ours\ can be further improved, as expected (\S \ref{SEC:SSIS}).

 \item
Since instance masks can provide much more precise %
localization than boxes, %
we envision
that \Ours\ can be used in many downstream tasks to boost their performance without  extra effort of annotating ground-truth masks.
For example, we can obtain text masks using \Ours (see \S \ref{SEC:CHAR}), which often help text recognition. \Ours\ can also help annotate the mask-level training data for the fully-supervised settings.

\end{itemize}

Instance segmentation has long been believed to be much more challenging to solve than bounding box detection.
Our strong performance of instance segmentation using only box supervision
shows that it may not necessarily be the case.

\subsection{Related Work}
\noindent\textbf{Box-supervised Semantic Segmentation.}
A few works                 %
attempted
to obtain
semantic masks using box annotations. For example, BoxSup~\cite{dai2015boxsup} uses the region proposals from MCG as the pseudo labels to train an FCN, and an iterative training algorithm is used to refine the estimated masks. The recent Box2Seg~\cite{kulharia12356box2seg} method employs the masks generated by GrabCut to supervise %
 training of the mask prediction model.
In addition, a per-class attention map is also predicted by the model to make the per-pixel cross entropy loss focus on foreground pixels and refine the segmentation boundaries. This method shows excellent performance on Pascal VOC~\cite{everingham2010Pascal}.
Authors of \cite{song2019box}
propose to use the unsupervised CRF~\cite{krahenbuhl2011efficient} to generate the segment proposals. Additionally, a class-wise filling rate loss to supervise the models
for
training, resulting in improved segmentation performance. One of the crucial steps in these methods is to employ the pseudo labels generated by unsupervised segmentation methods such as MCG~\cite{pont2016multiscale} or GrabCut~\cite{rother2004grabcut}. This is
because these method all rely on pixel-wise mask loss functions,
thus not being able to work without
mask annotations.
In this work, we %
remove the dependency on
pixel-wise mask losses, as a result, eliminating the region proposals.
Our new loss functions ensure that
 mask prediction can still be
 \textit{imperfectly}
 supervised %
 without using any mask annotations.

\noindent\textbf{Box-supervised Instance Segmentation.} In the context of deep learning, instance segmentation with box annotations has not explored too much yet. SDI~\cite{khoreva2017simple} might be the first instance segmentation framework with box annotations. Similar to the methods for semantic segmentation, SDI also relies on the region proposals generated by MCG. Then they make use of an iterative training procedure to further refine the segmentation results. Recently, BBTP~\cite{hsu2019weakly} formulates the box-supervised instance segmentation into a multiple instance learning (MIL) problem. BBTP is built on Mask R-CNN and samples the positive and negative bags according to the ROIs on CNN feature maps. In contrast, our method is built on ROI-free CondInst~\cite{tian2020conditional} and employs the proposed projection loss term to supervise the mask learning, eliminating the need for sampling. BBTP also makes use of the pairwise term. However,
their pairwise term is defined
on the set containing all neighboring pixel pairs
with the  oversimplified  assumption   of spatially neighboring pixel pairs
being encouraged to have the same label, inevitably introducing heavily noisy supervision. The crucial prior derived from proximal pixels' colors was not exploited in BBTP.
Our experiments  in Table~\ref{table:color_sim_threshold} show that, the heavily noisy
supervision can have a negative impact on accuracy.

As a result, we significantly outperform the mask AP of BBTP on COCO by an absolute 10\%.

\section{Approach}
\noindent\textbf{Conditional Convolutions for Instance Segmentation (CondInst)} Here, we briefly describe CondInst~\cite{tian2020conditional}. The main goal of CondInst is to solve instance segmentation in an RoI-free fully convolutional way. In CondInst, they believe that the most challenging piece in instance segmentation is that the prediction of each pixel varies according to the instance to be predicted. For example, when the model is predicting the mask for instance A, the pixels of instance B should be predicted as background. However, when the target instance is B, the pixels of B turn to be foreground. This poses the main challenge for the FCNs' application on instance segmentation because the traditional FCNs can only make deterministic prediction for each pixel.

\begin{figure}[t!]
\centering
\includegraphics[width=\linewidth]{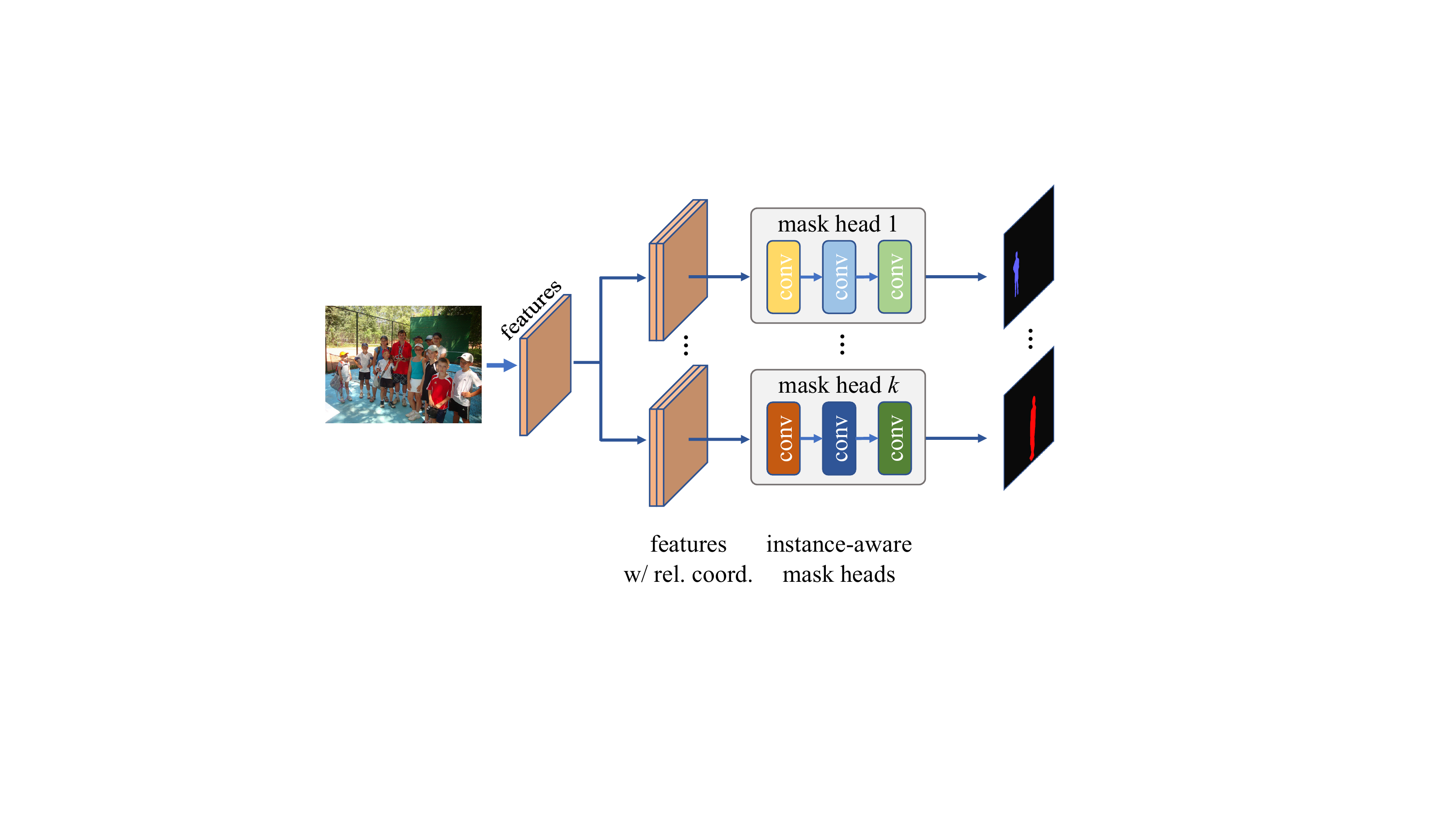}
   \caption{
   \textbf{Illustration of CondInst~\cite{tian2020conditional}.}
   CondInst employs the instance-aware dynamically-generated mask heads to obtain the full-image instance masks. We refer readers to \cite{tian2020conditional} for more details.}
\label{fig:condinst}

\end{figure}

CondInst \cite{tian2020conditional} proposes to employ dynamic filters~\cite{jia2016dynamic} to address the above issues. Instead of using a fixed mask head as in Mask R-CNN, according to the instance to be predicted, CondInst dynamically adapts the weights of the mask heads, as shown in Fig.~\ref{fig:condinst}, thus bypassing the above issue. With the instance-aware mask heads, CondInst can obtain the mask of each instance in the fashion of FCNs, eliminating the RoI operations. Notably, CondInst can generate the full-image instance masks, not only the masks within RoIs as in Mask R-CNN, as shown in Fig.~\ref{fig:condinst}. \textit{This has played a crucial role in the box-supervised settings}. Also, the mask head only predicts class-agnostic masks. The instance's category is determined by the detector's classification branch.

\subsection{Projection and Pairwise Affinity Mask Loss %
}
\noindent\textbf{Projection loss term.} As mentioned before, the first term supervises the horizontal and vertical projections of the predicted mask using the ground-truth box annotation, which ensures that the tightest box covering the predicted mask matches the ground-truth box. Formally, let $\vec{b} \in \{0, 1\}^{H \times W}$ be the mask generated by assigning $1$ to the locations in the ground-truth box and $0$ otherwise, as shown in Fig.~\ref{fig:vis_loss_terms} (top-right). Then we have
\begin{equation}
\begin{aligned}
    \Proj_x(\vb) = \vl_x,~~\Proj_y(\vb) = \vl_y,
\end{aligned}
\end{equation}
where $\Proj_x\colon \R^{H \times W} \to \R^W$ and $\Proj_y\colon \R^{H \times W} \to \R^H$ indicate that projecting the mask onto $x$-axis and $y$-axis, respectively. $\vl_x \in \{0, 1\}^W$ denotes the 1-D segmentation mask on $ x $-axis and the same applies to $\vl_y$.
The process of projection is illustrated in Fig.~\ref{fig:vis_loss_terms} (top row).

The projection operation can be implemented by a $\max$ operation along with each axis. Formally, we define
\begin{equation}
\begin{aligned}
    \Proj_x(\vb) = \max_y(\vb) = \vl_x, \\
    \Proj_y(\vb) = \max_x(\vb) = \vl_y,
\end{aligned}
\label{eq:proj}
\end{equation}
where $\max_y$ and $\max_x$ are the $\max$ operations along with $y$-axis and $x$-axis, respectively.

Let $\tilde{\vm} \in (0, 1)^{H \times W}$ be the network predictions for the instance mask, which can be viewed as the probabilities being foreground (\ie, the label is $1$).
We apply the same projection operations of Eq.~(\ref{eq:proj}) to the mask predictions and obtain the corresponding projections $\tilde\vl_x$ and $\tilde\vl_y$.
We then compute the loss between the projections of the ground-truth box and the predicted mask.
The projection loss term is defined as:
\begin{equation}
\begin{aligned}
    L_{proj} &= L(\Proj_x(\tilde{\vm}), \Proj_x(\vb)) + L(\Proj_y(\tilde{\vm}), \Proj_y(\vb)) \\
    &= L(\max_y(\tilde{\vm}), \max_y(\vb)) + L(\max_x(\tilde{\vm}), \max_x(\vb)) \\
    &= L(\tilde\vl_x, \vl_x) + L(\tilde\vl_y, \vl_y),
\end{aligned}
\end{equation}
where $L(\cdot, \cdot)$ is the Dice loss as in CondInst\footnote{One can also use the cross-entropy loss here.}. Note that \textit{all the operations in the last equation are (sub-)differentiable}.
This loss function is applied to all the instances in a training image and the final loss is their average. As shown in our experiments, by using this projection loss term, we can already obtain decent instance segmentation results without using any mask annotations.

\noindent\textbf{Pairwise affinity loss term.} In almost all instance segmentation frameworks such as Mask R-CNN and CondInst, they supervise the predicted masks in a per-pixel fashion.
The pixelwise supervision becomes unavailable if we do not have the mask annotations. Here, we attempt to supervise the mask in a \textit{pairwise} way, and we will show this supervision can be \textit{partially available} even if we do not have any mask annotations.

Now, assume we have the ground-truth masks. Consider an undirected graph $G = (V, E)$ built on an image, where $V$ is the set of the pixels in the image, and $E$ is the set of the edges. Each pixel is connected with its $K \times K - 1$ neighbours (the dilation trick may be applied), as shown in Fig.~\ref{fig:vis_loss_terms} (bottom left). Then we define $y_{e} \in \{0, 1\}$ be the label for the edge $e$, where $y_e = 1$ means the two pixels linked by the edge have the same ground-truth label and $y_e = 0$ means their labels are different. Let pixels $(i, j)$ and $(l, k)$ be the two endpoints of the edge $e$. The network prediction $\tilde{\vm}_{i, j} \in (0, 1)$ can be viewed as the probability of pixel $(i, j)$ being foreground. Then the probability of $y_e = 1$ is
\begin{equation}
\begin{aligned}
     P(y_e = 1) = \tilde{\vm}_{i, j}
     \cdot
     \tilde{\vm}_{k, l} + (1 - \tilde{\vm}_{i, j})
     \cdot
     (1 - \tilde{\vm}_{k, l}),
\end{aligned}
\end{equation}
and $P(y_e = 0) = 1 - P(y_e = 1)$. By convention, the probability distribution from the network prediction can be trained with the binary cross entropy (BCE) loss. %
Thus, the loss function is
\begin{equation}\label{eq:pairwise_loss}
\begin{aligned}
L_{pairwise} = -\frac{1}{N}\sum_{e \in E_{in}}y_{e}\log P(y_e = 1) \\
 +
 (1 - y_{e})\log P(y_e = 0),
\end{aligned}
\end{equation}
where $E_{in}$ is the set of the edges containing at least one pixel in the box. Using $E_{in}$ instead of $E$ here can prevent the loss from being dominated by a large number of the pixels outside the box. $N$ is the number of the edges in $E_{in}$.

If only the pairwise loss is used to supervise the mask learning (in the fully-supervised setting), ideally, two possible solutions may be obtained. The first one is the same as the ground-truth mask $\vm$, which is desirable. The second solution is the inverse $1 - \vm$. Fortunately, the second solution can be easily eliminated as long as we have a resolved label for any pixel. This can be achieved by the projection loss term because it ensures that the pixels outside the box is background. Note that the edges in $E_{in}$ still involve some pixels outside the box, which are of great importance to help the model get rid of the undesirable solutions. Overall, the total loss for mask learning can be formulated as
\begin{equation}\label{eq:loss_mask}
\begin{aligned}
L_{mask} = L_{proj} + L_{pairwise}.
\end{aligned}
\end{equation}
We will show in experiments that the redesigned mask loss can have similar performance to the original pixelwise one in the fully-supervised settings.

\subsection{Learning without Mask Annotations}
\begin{figure}[t!]
\centering
\includegraphics[width=0.997898\linewidth]{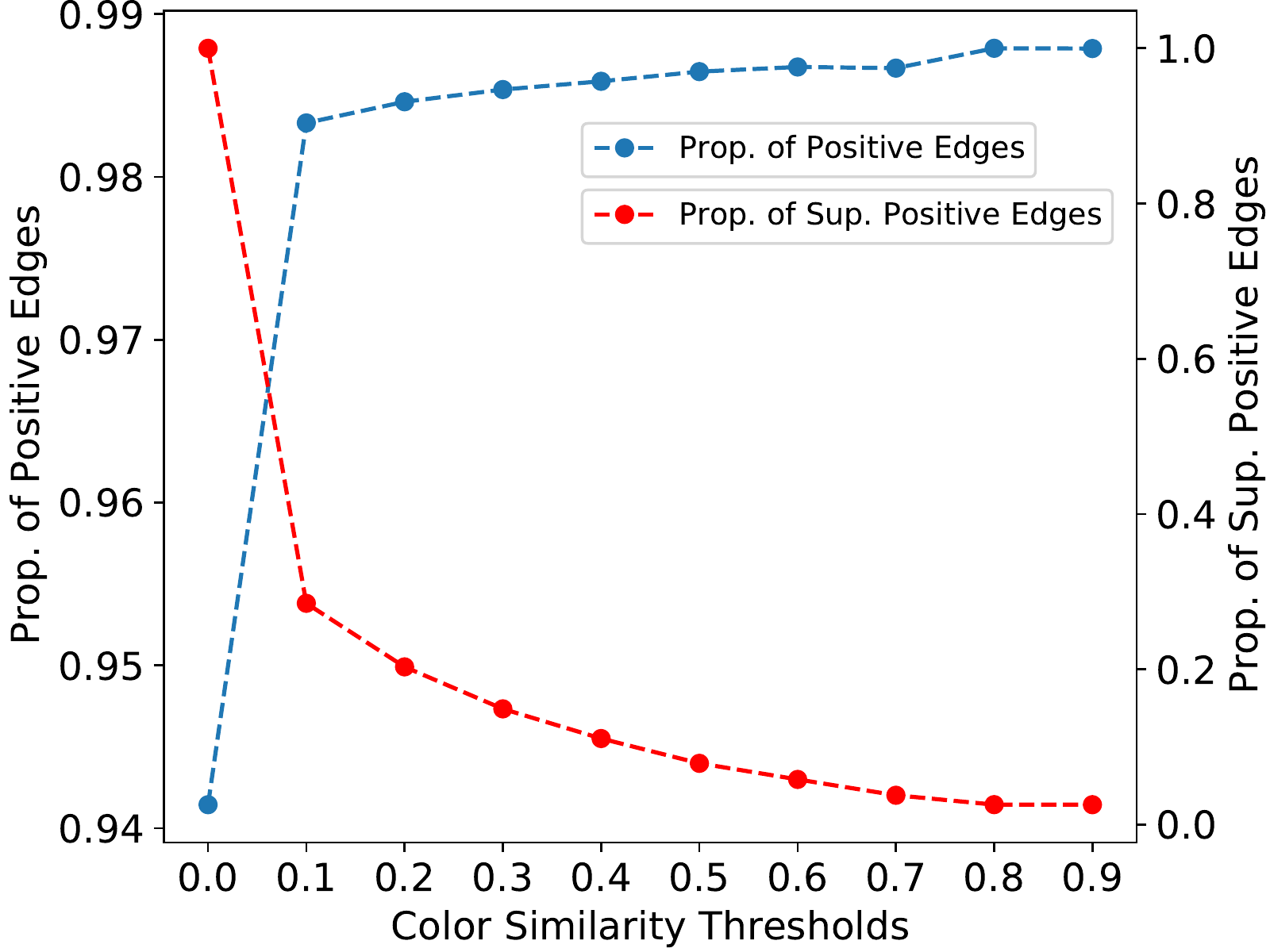}
\caption{\textbf{The relationship between the edges' labels and the color similarity thresholds.} `blue curve': the proportion of the positive edges in the edges with color similarity above the threshold. `red curve': the proportion of the supervised positive edges in all the positive edges. The number of positive edges are computed with the ground-truth masks of the COCO \texttt{val2017} split.}
\label{fig:color_sim_and_pos_edges}

\end{figure}
Thus far, we have shown that we can employ Eq.~(\ref{eq:loss_mask}) to train the mask. In Eq.~(\ref{eq:loss_mask}), the first term $L_{proj}$ is always valid no matter we have box or mask annotations. At first glance, the second term $L_{pairwise}$ still requires the mask annotations to compute the edge's label $y_e$. However, an important observation is that if two pixels have similar colors, they are very likely to have the same labels as well (\ie, the corresponding edge's label is 1). Thus, we can determine a color similarity threshold $\tau$ such that the edge's label is 1 with a high probability if its color similarity is above $\tau$. Formally, the color similarity is defined as
\begin{equation}\label{eq:color_sim}
\begin{aligned}
S_e = S(\vc_{i, j}, \vc_{l, k}) = \exp\left(-\frac{||\vc_{i, j} - \vc_{l, k}||}{\theta}\right),
\end{aligned}
\end{equation}
where $S_e$ be the color similarity of the edge $e$, and $\vc_{i, j}$ and $\vc_{l, k}$ are, respectively, the color vectors of the two pixels $(i, j)$ and $(l, k)$ linked by the edge. Here we use the LAB color space as it is closer to human perception. $\theta$ is a hyper-parameter, being $2$ in this work.

Then we can compute the pairwise loss for these confident edges only and discard the edges with agnostic labels. As a result, the pairwise loss becomes
\begin{equation}\label{eq:partial_pairwise_mask_loss}
\begin{aligned}
L_{pairwise} = -\frac{1}{N}\sum_{e \in E_{in}}\mathbbm{1}_{\{S_e \geq \tau\}}\log P(y_e = 1),
\end{aligned}
\end{equation}
where $\mathbbm{1}_{\{S_e \geq \tau\}}$ is the indicator function, being 1 if $S_e \geq \tau$ and 0 otherwise. Eq.~(\ref{eq:partial_pairwise_mask_loss}) only involves the term in Eq.~(\ref{eq:pairwise_loss}) for positive edges because we can only infer that the label of $e$ is 1 if $S_e \geq \tau$; but the label is agnostic if $S_e < \tau$.

We compute the proportion of the positive edges in the edges with $S_e \geq \tau$ with mask annotations on the COCO \texttt{val2017} split. Fig.~\ref{fig:color_sim_and_pos_edges} (blue curve) shows that the change of the proportion of the positive edges in the edges with $S_e \geq \tau$ as the threshold $\tau$ increases. As shown in figure, if the threshold is 0.1, more than $98\%$ of the edges are positive. The proportion can be further improved if we continue to increase $\tau$, but an overlarge threshold would reduce the number of the supervised edges (red curve in Fig.~\ref{fig:color_sim_and_pos_edges}). We need to trade off between them.

As we only have positive labels in Eq.~(\ref{eq:partial_pairwise_mask_loss}), one may note this would result in two possible trivial solutions, \ie, the masks of all the pixels being 0 or 1. However, the masks with all pixels being 0 do not meet the projection term; and the masks of all pixels being 1 almost never appear since the pairwise term encourages the pixels near the box boundaries to be negative if their colors are similar to that of the negative pixels outside the box.

\section{Experiments}
\begin{table*}
\setlength{\tabcolsep}{5pt}
\centering
\small
\subfloat[Varying the color similarity threshold $\tau$. \label{table:color_sim_threshold}]{
    \begin{tabular}{ l|c|c|c|c c|c c c }
     & prop. & AP & AP$_{50}$ & AP$_{75}$ & AP$_{S}$ & AP$_{M}$ & AP$_{L}$ \\
    \Xhline{2\arrayrulewidth}
    fully-sup. & - & 35.4 & 55.9 & 37.6 & 17.0 & 38.8 & 50.7 \\
    \hline
    $\tau = 0$ & 94.1\% & 9.4 & 30.3 & 3.3 & 7.6 & 10.3 & 11.4 \\
    $\tau = 0.1$ & 98.3\% & \textbf{30.7} & 52.2 & \textbf{31.1} & 13.8 & \textbf{33.1} & \textbf{45.7} \\
    $\tau = 0.2$ & \textbf{98.4\%} & 30.6 & \textbf{52.6} & 30.9 & \textbf{13.9} & 32.8 & 45.5 \\
    \end{tabular}
}
\hspace{0.5em}
\subfloat[Varying the size and dilation of the local patches (with $\tau = 0.1$). \label{table:varying_kernel}]{
    \begin{tabular}{ c|c|c|c c|c c c }
    size & dilation & AP & AP$_{50}$ & AP$_{75}$ & AP$_{S}$ & AP$_{M}$ & AP$_{L}$ \\
    \Xhline{2\arrayrulewidth}
    3 & 1 & 29.7 & 52.0 & 29.6 & 13.4 & 32.3 & 44.4 \\
    3 & 2 & \textbf{30.7} & 52.2 & \textbf{31.1} & \textbf{13.8} & \textbf{33.1} & \textbf{45.7} \\
    5 & 1 & 30.5 & \textbf{52.3} & 30.7 & 13.7 & 33.0 & \textbf{45.7} \\
    5 & 2 & 29.9 & 51.9 & 30.0 & 13.8 & 32.1 & 45.0 \\
    \end{tabular}
}
\caption{\textbf{Ablation study %
on
the hyper-parameters} in the proposed mask loss on the COCO  \texttt{val2017} split. ``prop." is the proportion of the positive edges in the edges with $S_e \geq \tau$. ``fully-sup.": fully-supervised results. As shown in Table~\ref{table:color_sim_threshold}, by using $\tau = 0.1$, \Ours\ can achieve 30.7 mask AP with only box annotations, which is close the fully-supervised mask AP (35.4\%) and significantly better localization precision than boxes (10.6\% mask AP as shown in Table~\ref{table:loss_terms}). Table~\ref{table:varying_kernel} shows the experiment results by varying the neighbours of each pixel.}

\end{table*}

\begin{table}
\centering
\small
\begin{tabular}{ l|c|c c|c c c }
mask loss & AP & AP$_{50}$ & AP$_{75}$ & AP$_{S}$ & AP$_{M}$ & AP$_{L}$ \\
\Xhline{2\arrayrulewidth}
Dice loss & 35.6 & 56.3 & 37.8 & 16.9 & 38.9 & 51.0 \\
proposed & 35.4 & 55.9 & 37.6 & 17.0 & 38.8 & 50.7 \\
\end{tabular}
\caption{\textbf{The projection and pairwise affinity mask loss vs.\  the original pixelwise one} in the fully-supervised settings. As
we can see here,
they %
attain very
similar mask AP on the COCO split \texttt{val2017}.}
\label{table:pixel_vs_pair}
\end{table}

We conduct %
experiments on COCO~\cite{lin2014microsoft} and Pascal VOC~\cite{everingham2010Pascal}. For COCO, the models are trained with \texttt{train2017} (115K images) and the ablation experiments are evaluated on \texttt{val2017} (5K images). Unless specified, \textit{only box annotations} are used during training. Our main results are reported on \texttt{test}-\texttt{dev}. For Pascal VOC, following previous works~\cite{hsu2019weakly, khoreva2017simple}, we train the models on the augmented Pascal VOC 2012 dataset~\cite{hariharan2011semantic} with 10, 582 training images, and evaluate them on Pascal VOC 2012 \texttt{val} split with 1, 449 images.

\subsection{Implementation Details}
Our proposed method only requires to change the mask loss in CondInst. Other training and testing details are kept as similar as possible to the original CondInst. On COCO, unless specified, we use the following training details. The models are trained for 90K iterations with batch size 16 on 8 V100 GPUs (2 images per GPU). The initial learning rate is set to 0.01 and reduced by a factor of 10 at step 60K and 80K, respectively. ResNet-50~\cite{he2016deep} is used as the backbone and is initialized with the ImageNet~\cite{deng2009imagenet} pre-training weights. The newly added layers are initialized as in ~\cite{tian2020conditional}. We use exactly the same data augmentation as in CondInst (\eg, left-right flipping and multi-scale training). For the dynamic mask heads, we use 3 conv.\ layers as in CondInst, but we increase the channels of the mask heads from 8 to 16, which can result in better performance with negligible extra computational overheads, and the compared baselines are adjusted accordingly. Following CondInst, the output mask is up-sampled to $\nicefrac{1}{4}$ resolution of the input image. For the pairwise loss term, we compute the pairwise relationship within $3 \times 3$ patches with the dilation rate being 2. On Pascal VOC, following \cite{hsu2019weakly}, we use batch size 8 and the number of iterations is 20K. The learning rate is reduced by a factor of 10 at step 15K. Only left-right flipping is used as the data augmentation during training. Other settings are the same as on COCO. The inference is the same as the original CondInst on both benchmarks. The performance is evaluated with the COCO-style mask AP.

\begin{table}[t!]
\setlength{\tabcolsep}{4pt}
\centering
\small
\begin{tabular}{ c|c|c|c c|c c c }
$L_{proj}$ & $L_{pairwise}$ & AP & AP$_{50}$ & AP$_{75}$ & AP$_{S}$ & AP$_{M}$ & AP$_{L}$ \\
\Xhline{2\arrayrulewidth}
\multicolumn{2}{c|}{box mask} & 10.6 & 32.2 & 4.6 & 5.7 & 11.3 & 15.6 \\
\hline
\checkmark & & 21.2 & 45.2 & 17.7 & 10.0 & 21.4 & 32.5 \\
\checkmark & \checkmark & \textbf{30.7} & \textbf{52.2} & \textbf{31.1} & \textbf{13.8} & \textbf{33.1} & \textbf{45.7} \\
\end{tabular}
\caption{\textbf{The mask AP on COCO \texttt{val2017}} by applying the different loss terms. ``box mask": using the masks generated by boxes. If both terms are not used, the model can only provide the box-level localization precision (10.6\% mask AP).}
\label{table:loss_terms}

\end{table}

\begin{table*}[t!]
	\centering
    \small
	\begin{tabular}{l| l |c|c|c|cc|ccc}
		method & backbone & aug. & sched. & AP & AP$_{50}$ & AP$_{75}$ & AP$_{S}$ & AP$_{M}$ & AP$_{L}$ \\
		\Xhline{2\arrayrulewidth}
	\hspace{-.3025cm}	\textit{fully supervised methods}: & & & & & & & \\
		Mask R-CNN~\cite{he2017mask} & ResNet-50-FPN & \checkmark & $3\times$ & 37.5 & 59.3 & 40.2 & 21.1 & 39.6 & 48.3 \\
		CondInst \cite{tian2020conditional} & ResNet-50-FPN & \checkmark & $3\times$ & 37.8 & 59.1 & 40.5 & 21.0 & 40.3 & 48.7 \\
		Mask R-CNN & ResNet-101-FPN & \checkmark & $3\times$ & 38.8 & 60.9 & 41.9 & 21.8 & 41.4 & 50.5 \\
		YOLACT-700~\cite{yolact} & ResNet-101-FPN & \checkmark & $4.5\times$ & 31.2 & 50.6 & 32.8 & 12.1 & 33.3 & 47.1 \\
		PolarMask \cite{xie2020polarmask} & ResNet-101-FPN & \checkmark & $2\times$ & 32.1 & 53.7 & 33.1 & 14.7 & 33.8 & 45.3 \\
		CondInst & ResNet-101-FPN & \checkmark & $3\times$ & 39.1 & 60.9 & 42.0 & 21.5 & 41.7 & 50.9 \\
		\hline
\hspace{-.3025cm} \textit{box-supervised methods}: & & & & & & & \\
		BBTP$^\dag$~\cite{hsu2019weakly} (prev.\ best) & ResNet-101-FPN & & $1\times$ & 21.1 & 45.5 & 17.2 & 11.2 & 22.0 & 29.8 \\
		\textbf{\Ours\!$^\dag$} & ResNet-101-FPN & & $1\times$ & 31.6 & 54.0 & 31.9 & 13.9 & 34.2 & 48.2 \\
		\textbf{\Ours} & ResNet-50-FPN & \checkmark & $3\times$ & 32.1 & 55.1 & 32.4 & 15.6 & 34.3 & 43.5 \\
		\textbf{\Ours} & ResNet-101-FPN & \checkmark & $1\times$ & 32.5 & 55.3 & 33.0 & 15.6 & 35.1 & 44.1 \\
		\textbf{\Ours} & ResNet-101-FPN & \checkmark & $3\times$ & 33.2 & 56.5 & 33.6 & 16.2 & 35.3 & 45.1 \\
		\textbf{\Ours} & ResNet-101-BiFPN~\cite{EfficientDetTanPL20} & \checkmark & $3\times$ & 33.9 & 57.7 & 34.5 & 16.5 & 36.1 & 46.6 \\
		\textbf{\Ours} & ResNet-DCN-101-BiFPN~\cite{zhu2019deformable} & \checkmark & $3\times$ & \textbf{35.0} & \textbf{59.3} & \textbf{35.6} & \textbf{17.1} & \textbf{37.2} & \textbf{48.9} \\
	\end{tabular}
	\caption{\textbf{Comparisons with state-of-the-art methods} on the COCO \texttt{test}-\texttt{dev} split. ``$\dag$"
	means that
	the results are on the COCO \texttt{val2017} split. BBTP only reported the results on the \texttt{val2017} split.
	Our \Ours\ outperforms the previous best reported mask AP by over absolute 10\% mask AP.
	Ours even outperforms two recent fully supervised methods, YOLACT and PolarMask, and is close to state-of-the-art fully-supervised results. `DCN': deformable convolutions~~\cite{zhu2019deformable}. `1$\times$' means 90K iterations.
	}
	\label{table:comparisons_state_of_the_art_methods}

\end{table*}

\subsection{Projection and Pairwise Affinity Loss for Mask Learning}
We first demonstrate that the redesigned mask loss can have similar performance to the original pixelwise mask loss in the fully-supervised settings. The experiments are conducted on COCO. We replace the original Dice loss for mask training in CondInst with the proposed one, and keep other settings exactly the same. As shown in Table~\ref{table:pixel_vs_pair}, the proposed mask loss can have similar performance (35.4\% vs.\  35.6\% mask AP), which suggests that using the proposed loss for mask learning is feasible.

\subsection{Box-supervised Instance Segmentation}
The key advantage of the proposed mask loss is that it can still supervise the predicted masks with only box annotations. We confirm this here and conduct experiments to investigate the hyper-parameters in the proposed mask loss.

\noindent\textbf{Varying the threshold of color similarity.} As mentioned before, we use a color similarity threshold $\tau$ to determine the edges that will be used to compute the pairwise loss. Here, we conduct experiments by varying $\tau$. When $\tau = 0$, all of the edges defined by the size of neighborhood are considered positive and used in the loss computation, as shown in Eq.~(\ref{eq:partial_pairwise_mask_loss}).
As shown in Table~\ref{table:color_sim_threshold}, in this case, 94.1\% of the edges are truly positive, and $\sim$6\% of the edges are actually negative. Thus, the loss computation would introduce $\sim$6\% noisy labels and all of the truly negative edges are wrongly labelled positive. Unsurprisingly, this experiment yields a trivial solution with poor performance (9.4\% mask AP) that almost all pixels in the box are predicted as foreground. If we increase $\tau$ to 0.1, the proportion of the truly positive edges are improved to 98.3\%, and only less than $\sim$2\% of the edges are wrongly labelled. As a result, the model can yield high-quality instance masks, achieving 30.7\% mask AP (vs.\  fully-supervised counterpart 35.4\%). This result is even better than that of some fully-supervised methods such as YOLACT and PolarMask. Some qualitative results are shown in Fig.~\ref{fig:vis_r101_3x}. If we further increase the threshold $\tau$ to 0.2, the performance slightly drops to 30.6\% mask AP. This might be because the number of the supervised positive edges decreases as we increase the threshold, as shown in Fig.~\ref{fig:color_sim_and_pos_edges}.

\noindent\textbf{Varying the neighborhood  of the pixels.} We conduct experiments with the different neighbours for each pixel. The size (\ie, $K$) defines how many surrounding pixels of each pixel are used to compute the pairwise loss with the pixel. Additionally, we may use the dilation trick to enlarge the scope (as in dilated convolutions). As shown in Table~\ref{table:varying_kernel}, by increasing the size from $3\times3$ to $5\times5$, the performance is boosted from 29.7\% to 30.5\%. This suggests that a relatively long-distance pairwise relationship is important to the final performance. However, using $5\times5$ requires more computational overheads in the training. Thus, we apply the dilation rate 2 to the $3 \times 3$ patches. This can capture the long-distance relationship without increasing the computational overheads and achieves similar performance (30.7\%). The performance cannot be further improved by applying the dilation trick to the $5\times5$ patches because the assumption, two pixels with similar colors probably have the same label, might not hold if the two pixels are far from each other.

\noindent\textbf{The contribution of each loss term.} Table~\ref{table:loss_terms} shows the contribution of each loss term. Even if only the first projection term is used, we can also achieve decent performance (21.2\% mask AP), which can already provide much higher localization precision than boxes (10.6\% mask AP). By further using the proposed pairwise term, high-quality instance masks can be obtained and the performance is largely improved to 30.7\%.

\subsection{Comparisons with State-of-the-art}
We compare \Ours\ with state-of-the-art fully/box supervised instance segmentation methods on the COCO dataset. As shown in Table~\ref{table:comparisons_state_of_the_art_methods}, with the same backbone and training settings, \Ours\ significantly surpasses the previous best reported result~\cite{hsu2019weakly} by absolute 10.5\% mask AP (\eg, from 21.1\% to 31.6\%).
\textit{\Ours, without using any mask annotations, performs even better than some recent fully-supervised methods such as PolarMask~\cite{xie2020polarmask} and YOLACT~\cite{yolact}}, with the same backbones as well as similar training and testing settings (32.5\% with R-101 $1\times$ vs.\ PolarMask 32.1\% R-101 $2\times$ and YOLACT-700 31.2\% R-101 $4.5\times$). \Ours\ also %
demonstrates
competitive %
performance with %
top-performing
fully-supervised instance segmentation methods. For example, with the same backbone ResNet-50-FPN and $3\times$ training schedule, \Ours\ achieves 32.1\% mask AP (vs. 37.8\% of the fully-supervised CondInst~\cite{tian2020conditional}). With BiFPN~\cite{EfficientDetTanPL20} and DCN~\cite{zhu2019deformable}, the performance can be further boosted to 35.0\% mask AP. Some qualitative results are shown in Fig.~\ref{fig:vis_r101_3x}. The excellent performance shows that \Ours\ dramatically
narrows the performance
gap between the fully supervised and box-supervised instance segmentation, and for the first time, the great potential of box-supervised instance segmentation is revealed.

\subsection{Experiments on Pascal VOC}
\begin{table}
\centering
\small
\begin{tabular}{ l|l|c|c c}
method & backbone & AP & AP$_{50}$ & AP$_{75}$ \\
\Xhline{2\arrayrulewidth}
GrabCut~\cite{rother2004grabcut} & ResNet-101 & 19.0 & 38.8 & 17.0 \\
SDI~\cite{khoreva2017simple} & VGG-16 & - & 44.8 & 16.3 \\
BBTP~\cite{hsu2019weakly} & ResNet-101 & 23.1 & 54.1 & 17.1 \\
BBTP w/ CRF & ResNet-101 & 27.5 & 59.1 & 21.9 \\
\hline
\textbf{\Ours} & ResNet-50 & 34.3 & 59.1 & 34.2 \\
\textbf{\Ours} & ResNet-101 & \textbf{36.5} & \textbf{61.4} & \textbf{37.0} \\
\end{tabular}
\caption{
\textbf{Results on Pascal VOC \texttt{val2012}}. %
\Ours\ surpasses previous methods by a large margin. Here, the GrabCut obtains the instance masks by taking as input the boxes generated by \Ours. Thus, the only difference between the GrabCut and \Ours\ is the way to obtain the masks.
Clearly,
\Ours\ %
achieves
significantly
improved mask AP, outperforming previous best by about 10\%.
}
\label{table:experiments_on_voc}

\end{table}
We also conduct experiments on Pascal VOC. As shown in Table~\ref{table:experiments_on_voc}, \Ours\ achieves state-of-the-art instance segmentation with only box annotations. With the same backbone and training settings, \Ours\ outperforms BBTP both in AP$_{50}$ and AP$_{75}$ by a large margin. Notably, the AP$_{75}$ is improved by more than relative 200\% (17.1\% vs.\ 37.0\% mask AP), which suggests \Ours\ can produce the masks of much higher quality. \Ours\ is even much better than the BBTP with CRF. Additionally, \Ours\ also performs much better than SDI~\cite{khoreva2017simple}. We also compare \Ours\ with the traditional unsupervised segmentation method GrabCut~\cite{rother2004grabcut}. In the experiment, GrabCut takes as input the bounding-boxes predicted by the ResNet-101 based FCOS in \Ours. Thus the only difference between \Ours\ and GrabCut is the way of obtaining instance masks. As shown in Table~\ref{table:experiments_on_voc}, \Ours\ is far better than GrabCut (19.0\% vs.\ 36.5\% mask AP). Moreover, \Ours\ is fully convolutional and can benefit from the highly-efficient GPUs, thus inferring tens of times faster than GrabCut.

\subsection{Extensions: Semi-supervised Instance Segmentation}\label{SEC:SSIS}
\begin{table}
\setlength{\tabcolsep}{4pt}
\centering
\small
\begin{tabular}{ c|c|c c c|c c c}
\multirow{2}{*}{$L_{proj}$} & \multirow{2}{*}{$L_{pairwise}$} & \multicolumn{3}{c|}{all 80 classes} & \multicolumn{3}{c}{60 unseen classes} \\
 & & AP & AP$_{50}$ & AP$_{75}$ & AP & AP$_{50}$ & AP$_{75}$ \\
\Xhline{2\arrayrulewidth}
& & 24.7 & 44.6 & 24.2 & 19.9 & 38.3 & 18.5 \\
\checkmark & & 31.8 & 52.5 & 33.2 & 29.7 & 49.3 & 31.0 \\
\checkmark & \checkmark & \textbf{32.5} & \textbf{53.0} & \textbf{34.0} & \textbf{30.9} & \textbf{50.1} & \textbf{32.4} \\
\hline
\multicolumn{2}{c|}{box supervised} & 30.7 & 52.2 & 31.1 & 29.6 & 49.7 & 30.4 \\
\end{tabular}
\caption{
\textbf{\Ours\ for semi-supervised instance segmentation}.
These models are trained with the 20 classes mask annotations and the other 60 classes (\ie, unseen classes) are only with box annotations.}
\label{table:semi_sup_voc_to_nonvoc}

\end{table}

\begin{table}[t!]
\setlength{\tabcolsep}{4pt}
\centering
\small
\begin{tabular}{ c|c|c c c|c c c}
\multirow{2}{*}{$L_{proj}$} & \multirow{2}{*}{$L_{pairwise}$} & \multicolumn{3}{c|}{all 80 classes} & \multicolumn{3}{c}{20 unseen classes} \\
 & & AP & AP$_{50}$ & AP$_{75}$ & AP & AP$_{50}$ & AP$_{75}$ \\
\Xhline{2\arrayrulewidth}
& & 32.1 & 51.6 & 33.9 & 25.5 & 45.5 & 25.1 \\
\checkmark & & 33.1 & 53.8 & 34.3 & 31.6 & 57.4 & 30.0 \\
\checkmark & \checkmark & \textbf{33.8} & \textbf{54.3} & \textbf{35.7} & \textbf{35.9} & \textbf{60.9} & \textbf{36.3} \\
\hline
\multicolumn{2}{c|}{box supervised} & 30.7 & 52.2 & 31.1 & 29.6 & 49.7 & 30.4 \\
\end{tabular}
\caption{\textbf{\Ours\ for semi-supervised instance segmentation}. The models are trained with the 60 classes mask annotations and other 20 classes (\ie, unseen classes) are only with box annotations.}
\label{table:semi_sup_non_voc_to_voc}

\end{table}

In this section, we show that our method can also help the model generalize to unseen categories in the semi-supervised setting where only partial classes have the mask annotations. Following previous works~\cite{Zhou2020ShapeProp, hu2018learning, kuo2019shapemask} in this setting, we conduct the experiments on the COCO dataset and split the $80$ classes in COCO into two groups -- 20 classes present in Pascal VOC and 60 classes not in Pascal VOC. Then the models are trained with the mask annotations of the classes in one group, and another group of classes only have the box annotations. The generalization ability is evaluated with the mask AP averaged over the group of classes without mask annotations (\ie, unseen classes).

We first train the model with the 20 classes mask annotations. As shown in Table~\ref{table:semi_sup_voc_to_nonvoc} (1st row), if our proposed loss terms are not used, where the mask loss is only computed for the instances with mask annotations and other instances are discarded during the mask learning, the model can only achieve 19.9\% mask AP on the unseen categories. This low performance suggests that the model is difficult to generalize to unseen classes. If we use the $L_{proj}$ term for the 60 classes without the mask annotations during training, as shown in the table (2nd row), the performance can be dramatically improved to 29.7\%. If we further apply the pairwise term $L_{pairwise}$, the performance can be boosted to 30.9\%. Moreover, compared to the setting only using the box annotations (last row in Table~\ref{table:semi_sup_voc_to_nonvoc}), the performance on the unseen classes is also improved from 29.6\% to 30.9\%, which suggests that the box-supervised model can benefit from the partial mask annotations. Additionally, the experimental results with the 60 classes masks are shown in Table~\ref{table:semi_sup_non_voc_to_voc}, and the same conclusions can be drawn.

\subsection{Extensions: Box-supervised Character Segmentation} \label{SEC:CHAR}

\begin{figure}[t!]
\centering
\includegraphics[width=\linewidth]{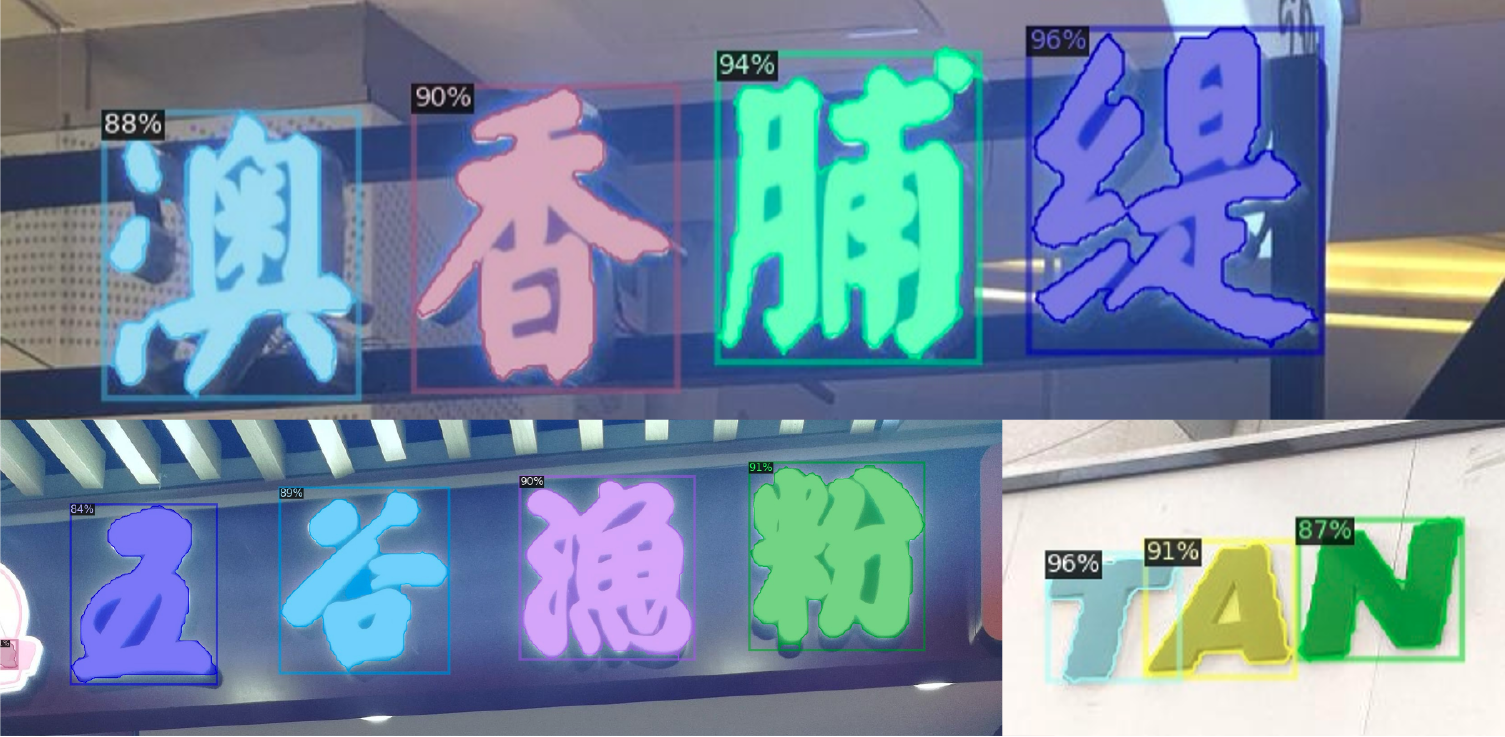}
   \caption{%
   \textbf{Character masks %
   predicted
   by \Ours.} %
   No
   mask annotations are used %
   for
   training.}
\label{fig:char_segmentation}

\end{figure}

 In order to
 demonstrate the generality of \Ours, we conduct experiments to obtain the character masks with character box annotations. Our experiments are conducted on the ICDAR 2019 ReCTS dataset ~\cite{zhang2019icdar}, which %
contains
20K training images and 5K testing images. These images are annotated with text-line and character-level boxes. We train our model with the character boxes. All the training settings are the same as that of COCO. Since we do not have mask annotations for the testing set, it is impossible to
report the mask AP. We instead show some qualitative results in Fig.~\ref{fig:char_segmentation},
demonstrating that
\Ours\ can obtain high-quality character masks.
The text masks might provide useful cues for
detecting and recognising text of arbitrary shapes.
We believe that the ability of \Ours generating
character masks automatically may inspire new applications on this task.

\section{Conclusions}

In this work, we have proposed \Ours\ that can achieve high-quality instance segmentation with only box annotations. The core idea of \Ours\ is to replace the original pixelwise mask loss with the proposed projection and pairwise affinity mask loss. With the proposed mask loss, we show excellent instance segmentation performance without using any mask annotations on COCO and Pascal VOC, significantly improving the state-of-the-art.

{\small
\bibliographystyle{ieee_fullname}
\bibliography{draft}
}

\end{document}